\documentclass[conference]{IEEEtran}
\IEEEoverridecommandlockouts
\usepackage{cite}
\usepackage{amsmath,amssymb,amsfonts}
\usepackage{algorithmic}
\usepackage{graphicx}
\usepackage{tabularx}
\usepackage{hyperref}
\hypersetup{
    colorlinks=true,
    linkcolor=blue,  % You can set the color to your preference
    linktoc=page
}
\usepackage{graphicx}
\usepackage{textcomp}
\usepackage{xcolor}
\def\BibTeX{{\rm B\kern-.05em{\sc i\kern-.025em b}\kern-.08em
    T\kern-.1667em\lower.7ex\hbox{E}\kern-.125emX}}
\begin{document}

\title{AINS: Affordable Indoor Navigation Solution via Line Color Identification Using Mono-Camera for Autonomous  Vehicles\\
\thanks{Identify applicable funding agency here. If none, delete this.}
}

\author{\IEEEauthorblockN{1\textsuperscript{st} \textbf{Nizamuddin Maitlo$^*$}}
\IEEEauthorblockA{\textit{Department of Computer Science} \\
\textit{Sukkur IBA University}\\
Sukkur, Pakistan \\
Corresponding Author: nizamuddin.cs@iba-suk.edu.pk
}
\and
\IEEEauthorblockN{2\textsuperscript{nd} Nooruddin Noonari}
\IEEEauthorblockA{\textit{Department of Computer Science} \\
\textit{IBA Community College, Naushahro Feroze}\\
Sindh, Pakistan\\
Corresponding Author: noor.cs2@yahoo.com}

\and
\IEEEauthorblockN{3\textsuperscript{rd} Kaleem Arshid}
\IEEEauthorblockA{\textit{DITEN} \\
\textit{University of Genova}\\
Genova, Italy\\
kaleem.arshid@edu.unige.it}

\and
\IEEEauthorblockN{4\textsuperscript{th} Naveed Ahmed}
\IEEEauthorblockA{\textit{IBA Community College Naushahro Feroze} \\
\textit{Sukkur IBA University}\\
Sindh, Pakistan\\
engr.naveed.pv@gmail.com}

\and
\IEEEauthorblockN{5\textsuperscript{th} Sathishkumar Duraisamy}
\IEEEauthorblockA{\textit{Department of Mechanical Engineering} \\
\textit{Kathir College of Engineering}\\
Tamilnadu, India.\\
sathizkumard@gmail.com}
}

\maketitle

\begin{abstract}
Recently, researchers have been exploring various ways to improve the effectiveness and efficiency of autonomous vehicles by researching new methods, especially for indoor scenarios. Autonomous Vehicles in indoor navigation systems possess many challenges especially the limited accuracy of GPS in indoor scenarios. Several, robust methods have been explored for autonomous vehicles in indoor scenarios to solve this problem, but the ineffectiveness of the proposed methods is the high deployment cost.

To address the above-mentioned problems we have presented A low-cost indoor navigation method for autonomous vehicles called Affordable Indoor Navigation Solution (AINS) which is based on based on Monocular Camera. Our proposed solution is mainly based on a mono camera without relying on various huge or power-inefficient sensors to find the path, such as range finders and other navigation sensors. Our proposed method shows that we can deploy autonomous vehicles' indoor navigation systems while taking into consideration the cost. We can observe that the results shown by our solution are better than existing solutions and we can reduce the estimated error and time consumption.

\end{abstract}


\section{Introduction}
With the advancement of computer vision technology now we can give the power of vision to machines and vehicles. The research on autonomous vehicles is becoming more prominent to replace the traditional way of driving to improve the quality of travel and reduce the chances of human error like fatigue, and tiredness. Automation is widely welcomed in all aspects of life to replace conventional transportation methods for industrial usage, self-driving vehicle indoor navigation system is a popular research topic in the field of automation \cite{b1}\cite{b2}. 

Despite the popularity of indoor navigation systems researchers are facing challenges while developing autonomous vehicle navigation methods due to the low precision of GPS. Researchers are exploring many methods and there is a potential area that can be further explored to enhance the robustness of the indoor navigation system for autonomous vehicles \cite{b3}\cite{b4}.

There are several methods implemented in self-driving vehicles to achieve the required task such as laser-based simultaneous localization and mapping (SLAM) and visual SLAM \cite{b5}. Both approaches come equipped with the pros and cons such as the cost of the inertial sensors-based navigation-based vehicle method \cite{b6}. This method is not feasible for those with limited budgets and who are looking to gain prosperity. There is a wide range of methods required to implement in order to navigate a vehicle in an indoor scenario, such as object tracking, angle prediction, steering module, path planning, etc.

There are lots of methods that already exist for indoor vehicle navigation depending on different kinds of hardware and software. These hardware and software components can be cheap and expensive, but the results of low-price components are below average. Ultrasonic or infrared sensors can be utilized for the navigation of autonomous vehicles but more than one sensor is required and a very specific environment to self-navigate. The environmental noise is a problem for these sensors because these sensors can be easily affected by noise resulting in the durability and performance of the sensors \cite{b7}\cite{b8}.

In \cite{b9}\cite{b10}, a method, which can track the moving object and map the environment using a sequence of images is introduced. Their method used the Kalman filter algorithm, which has to detect the feature points, but first whole space is mapped as well and the system initially starts with known features and then the upcoming objects are detected using certain features. On the other hand \cite{b11}, also presented a vehicle motion model using a sequence of images and three degrees of freedom IMU with a digital camera. The low-quality IMU can be affected easily by noise, which will tremendously affect vehicle motion.

To address the above-mentioned problems we have presented A low-cost indoor navigation method for autonomous vehicles called Affordable Indoor Navigation Solution (AINS) which is based on based on Monocular Camera. In this paper, we have proposed a robust method based on image processing in order to achieve indoor vehicle navigation. The popular method of using color lines to navigate a vehicle in the indoor scenario has been studied widely in the few past years. The vehicle can be navigated using several line methods but in our proposed system we will be using image processing methods such as color object tracking \cite{b12}. This method aims to keep the self-driving vehicle on the path and deliver the package after traveling a certain path. There are a few things that are important to achieve the desired goal. The vehicle has to be very stable and precise with its movements, proper rotation along the path, reduce environmental effects, and reach its destination at a very specific timeline.

Accordingly, the further article is organized as follows. The following section II briefly describes the literature review. Section III briefly describes the proposed methodology of the indoor navigation system of autonomous vehicles. Section IV results and discussion contains the results and outcomes of the proposed methods. In section V article is concluded at the end.

\section{Indoor Navigation System}
\begin{figure}[htbp]
\centering
\includegraphics[width=0.4\textwidth]{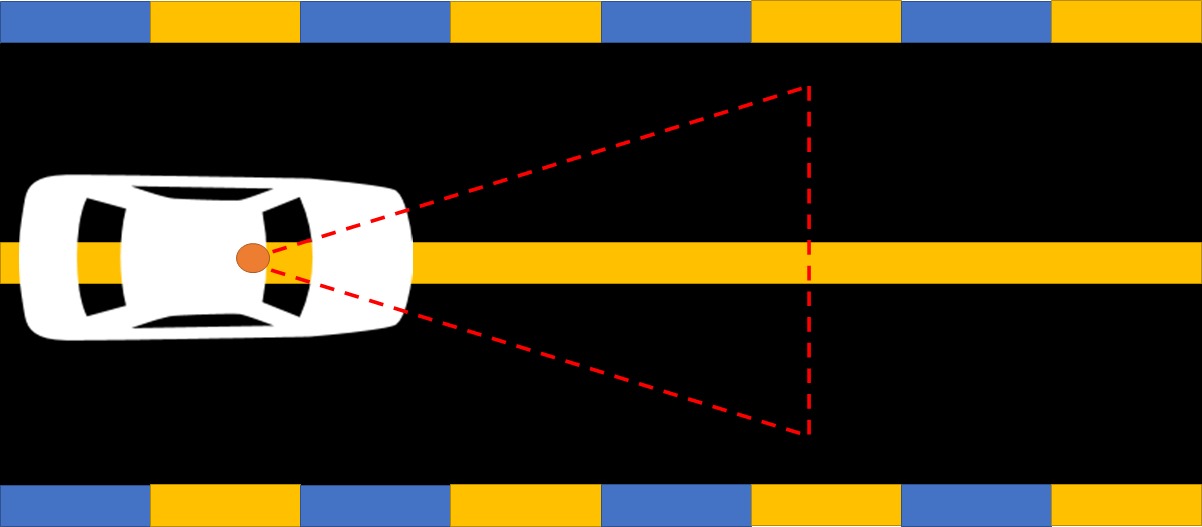} % Adjust the width as needed
\caption{Indoor Navigation System}
\label{pic01}
\end{figure}

In \autoref{pic01} indoor navigation system working mechanism is elaborated as you can see the camera mounted on the vehicle continuously scans the path to identify a line by processing the input stream and searching for the specified color line and once the line is identified the angle in the line is calculated to make precise moves by vehicle. One path is determined as the vehicle progresses forward.

\section{Literature Review}

Currently, researchers are exploring various ways to improve the effectiveness and efficiency of self-driving vehicles by proposing new techniques, specifically for indoor scenarios. Self-driving vehicles in indoor navigation systems face challenges such as the limited accuracy of GPS. These methods are as follows.

In this paper \cite{b13}, a pioneering approach to autonomous vehicle navigation in indoor settings are introduced, utilizing omni-directional vision and circular landmarks affixed to the ceiling. The use of upward-looking omni-directional images minimizes occlusion and noise, enhancing location estimation. Analytic formulas demonstrate that the circular landmark's perspective shape in the omni-directional image can be accurately approximated by an ellipse, enabling precise vehicle localization for effective navigation, as validated through simulations and real-world experiments.

This study \cite{b14}, advocates for smartphones as a promising solution to ubiquitous positioning issues, emphasizing the need for minimal impact on device battery life. Introducing SmartSLAM, a novel indoor positioning system, the research proposes an intelligent filtering approach that dynamically adapts sensor fusion algorithms based on system certainty levels. This innovation effectively reduces computational load, enhances positioning accuracy, prolongs battery life, and allocates CPU resources more efficiently for foreground processes.

The following research \cite{b15}, tackles the challenging problem of indoor positioning in GPS-denied environments, crucial for applications like augmented reality and autonomous systems. Introducing a novel tandem architecture of deep network-based systems, the approach leverages scene images obtained through photogrammetry, employing an EfficientNet-based CNN for scene classification and a MobileNet-based CNN for regression. The system achieves remarkable precision in both Cartesian position and quaternion information for camera localization, marking a significant advancement in indoor positioning technology.

This paper \cite{b16} addresses the surging applications of unmanned aerial vehicles (UAVs) across diverse domains, emphasizing the growing significance of indoor use cases such as rescue operations and warehouse inventory tracking. The study introduces a novel scheme employing deep neural networks, specifically a DCNN-GA architecture, for autonomous UAV navigation in indoor corridors, overcoming challenges of accurate localization and obstacle avoidance. Notably, the proposed approach utilizes genetic algorithms for hyperparameter tuning, demonstrating superior performance compared to state-of-the-art ImageNet models in terms of minimal loss and heightened efficiency.

This study \cite{b17} introduces a real-time monocular vision-based range measurement method tailored for Simultaneous Localization and Mapping (SLAM) in GPS-denied environments, specifically designed for an Autonomous Micro Aerial Vehicle (MAV) with limited payload capacity. The navigation strategy relies on corner-based feature points extracted from a monocular camera, presenting a solution for GPS-denied manmade environments. The experimental validation involves a case study mission demonstrating vision-based pathfinding through a conventional maze of corridors in a sizable building.

This study \cite{b18} proposed a unique monocular camera-based indoor navigation and ranging strategy for autonomous self-driving vehicles. They are utilizing the orthogonality of the indoor environment architecture by introducing a new technique using a mono camera for Visual-SLAM to estimate the range and state of the vehicle. The navigation approach presupposes an artificial indoor or indoor-like environment with an unknown layout, denied the GPS information, and characterized by features based on energy and straight architectural lines. 
The proposed algorithms were tested on a fully autonomous microaerial vehicle (MAV) equipped with advanced onboard image processing and SLAM capabilities. Results from the experiments indicate that the system's performance is constrained solely by the camera's capabilities and the environmental complexity.

This study \cite{b19} presents an autonomous integrated indoor navigation system for ground vehicles, leveraging a fusion of inertial sensors, LiDAR, wireless local area network (WLAN) signal strength, odometry, and predefined floor maps. Addressing the challenge of self-alignment and position initialization in the absence of GNSS, the system employs an extended Kalman filter for tilt angle estimation and a subimage matching algorithm for initial position and heading estimation using 2-D LiDAR scans. The approach is validated in a real office environment, demonstrating effective self-alignment, initialization, and submeter-level positioning accuracy.

This Study \cite{b20} introduces a real-time path-planning algorithm for indoor UAV navigation, specifically designed for contact inspection tasks in diverse environments. Utilizing only the point cloud of the building as input, the algorithm comprises a pre-processing step for segmentation and discretization, enabling swift execution of the real-time path planning algorithm. Demonstrated in various buildings, the method consistently achieves route calculations in 8–9 milliseconds, meeting execution time requirements and establishing its reliability for UAV route calculation in indoor settings.

\section{Methodology}
\subsection{Transitional Behaviour of Framework}
\begin{figure}[htbp]
\centering
\includegraphics[width=0.5\textwidth]{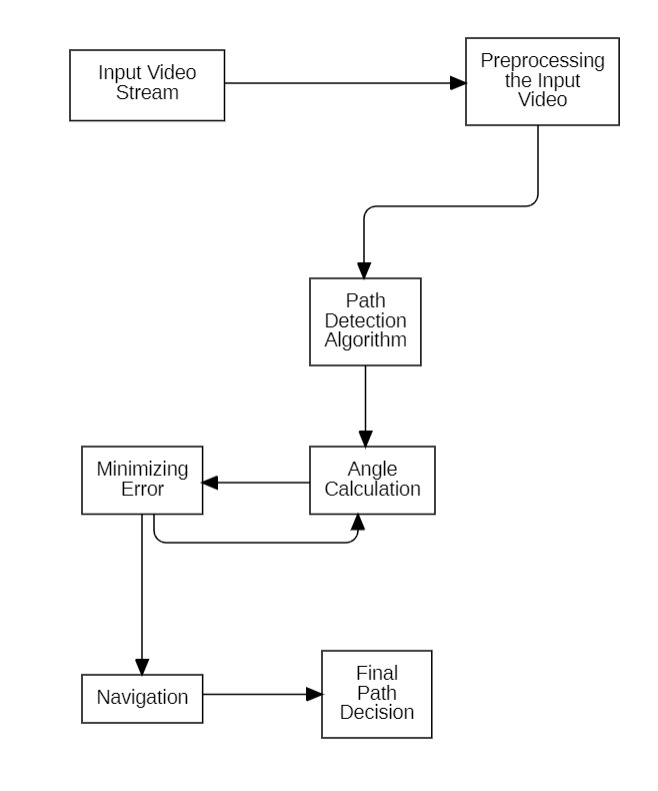} % Adjust the width as needed
\caption{Transitional Behaviour of Framework}
\label{pic02}
\end{figure}

The proposed framework AINS flow-work is shown in \autoref{pic02} which elaborates on the steps taken by the navigation method. Firstly the method takes an input video stream from the camera mounted on the vehicle. Secondly, the input video stream is further prepossessed and fed to the path detection method. Once the path color is detected the angle is calculated for possible turns in the path minimizing the possible turn errors. lastly, the navigation module takes place, and the final path is detected.

\subsection{Purposed Framework (AINS)}
\begin{figure}[htbp]
\centering
\includegraphics[width=0.5\textwidth]{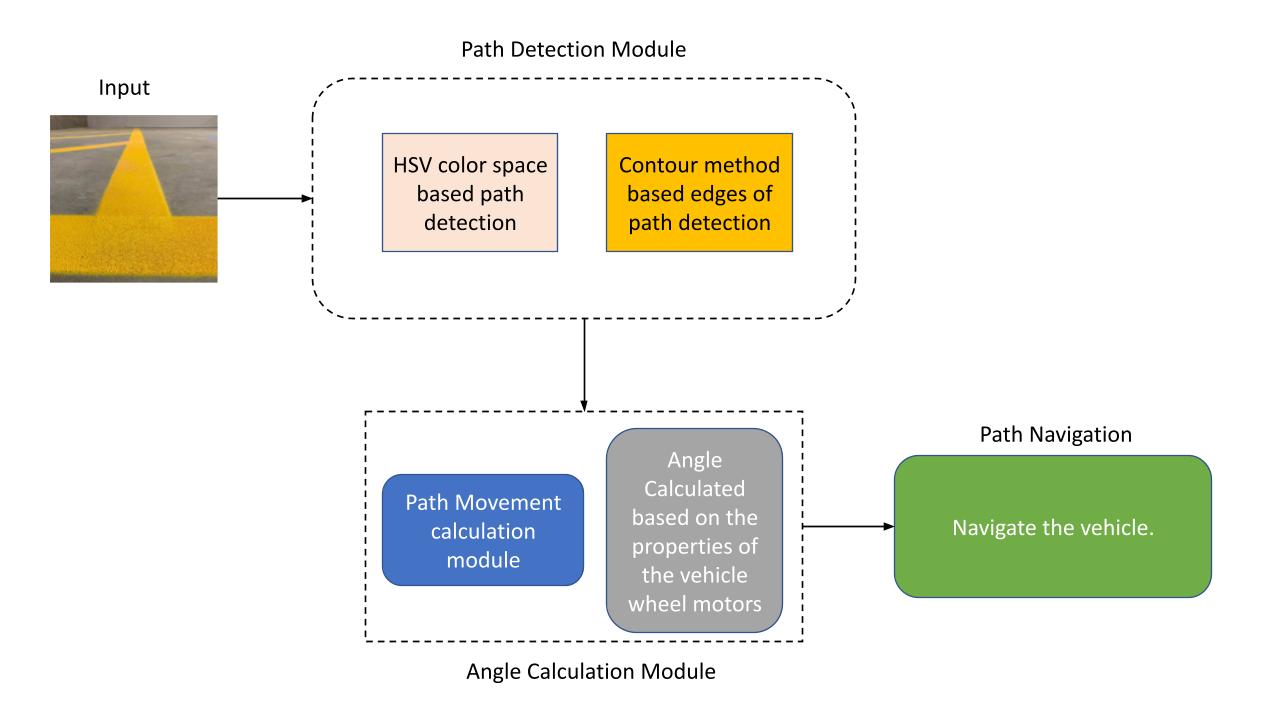} % Adjust the width as needed
\caption{The Proposed Framework}
\label{pic03}
\end{figure}

We have proposed a vision-based navigation system called Affordable Indoor Navigation Solution (AINS) for our prototype vehicles as shown in  
 \autoref{pic03}. In AINS fixed RGB camera is equipped with a vehicle. AINS consists of different components including motors, microcontroller units, and motion controllers. Our proposed AINS proved that with an RGB camera, we can automatically detect a path and then follow that path correctly. In the proposed framework, we have experimentally shown that our system detects path and navigate vehicle camera using the mono camera. In the AINS framework, an RGB camera is mounted to the front side of the vehicle facing the path. The resolution of the captured image is 640×480 pixels. This experiment is based on real-time performance, the vision application will detect the path and then the vehicle will run along the path to finish the maneuver.

\subsection{The Methodical Stages of Our Approach}
\subsubsection{Image Feature Extraction}

The proposed AINS framework has first and foremost step is the Image framework extraction, where we have used the Gaussian Filtering technique. We have used Gaussian filtering first to remove the noise from the image and then used the image for further process, the following \eqref{eq:01} is the Gaussian Filtering formula for Image Feature Extraction.

\begin{equation}
G(x, y) = \frac{12\pi^2}{e^{x^2 + \frac{y^2}{22}}}
\label{eq:01}
\end{equation}

In the Gaussian Filtering, x and x represent the point coordinates system and they behave as Gaussian distribution. These x and y point coordinates control the noise-removing process. After the de-noising step, we used the following \eqref{eq:02} to smooth the image.

\begin{equation}
f'(x, y) = G(x, y) * f(x, y)
\label{eq:02}
\end{equation}

Where:
\begin{itemize}
  \item $G(x, y)$ represents the Gaussian filter,
  \item $f(x, y)$ represents the original image, and
  \item $f'(x, y)$ represents the filtered image.
\end{itemize}

\subsubsection{Spotting Anomalies on the Path}

\begin{figure}[htbp]
\centering
\includegraphics[width=0.5\textwidth]{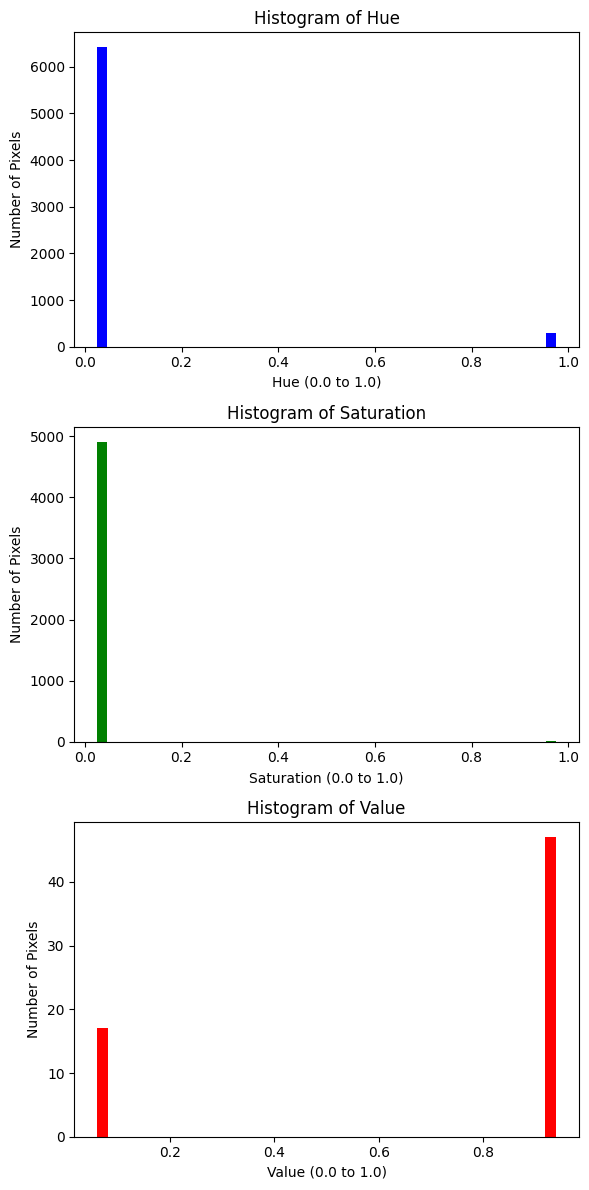} % Adjust the width as needed
\caption{Histogram of H,S,V}
\label{pic04}
\end{figure}

In our AINS proposed framework, we have performed an experiment based on HSV color space and further used it to identify the path colors. The camera first detects the image then it identifies the color in RGB space, after identifying the path colors, the image colors are transferred from RGB form to HSV form. Among other color models, the HSV color model is ideal for providing a more human way to describe color, because hue, saturation, and value components are closely related to how the eye perceives color. In the HSV space, hue represents the hue, saturation represents the grayscale of the color space, and value represents the brightness of the image [\cite{b21}\cite{b22}. Once image transformation is completed, our solution identifies all pixels from the HSV color space and we present those HSV pixels in the Histogram displayed in \autoref{pic04}. The graph shows that the distribution of the HSV pixels represents the all viable count of pixels in the path color to be tracked.

\begin{figure}[htbp]
\centering
\includegraphics[width=0.5\textwidth]{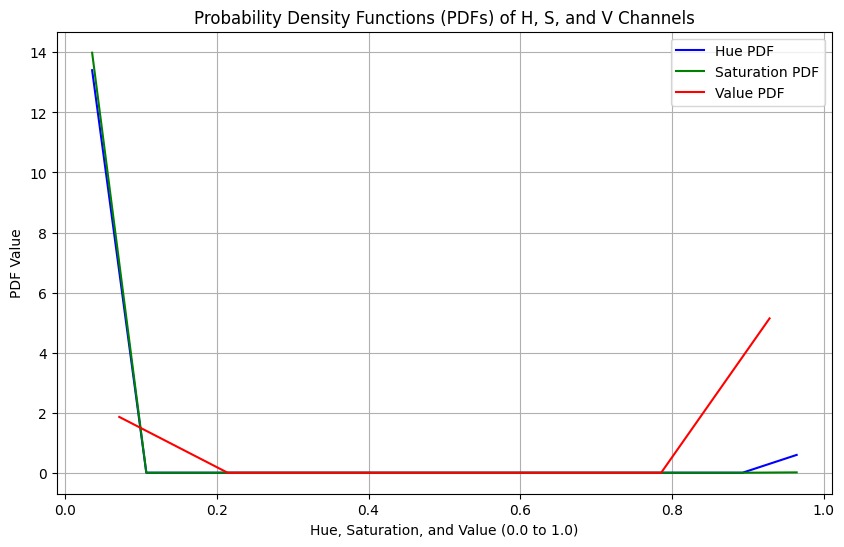} % Adjust the width as needed
\caption{PDF distribution of Pixels}
\label{pic05}
\end{figure}

Furthermore, for displaying pixels as colors space HSV as probability distribution we have used the probability density function (PDF), as PDF has mean and standard deviation so we have shown the distribution of each pixel in \autoref{pic05}. In this graph, the standard deviation is used to measure the upper and lower bounds of the pixel values, and also ignore outliers, and pixels that are not connected with paths.

%Tabl01
\begin{table}[htbp]
\centering
\caption{Ranges of HSV Values for Identifying Yellow Color}
\label{tab:hsv-yellow-color}
\begin{tabular}{|c|c|c|}
\hline
Criteria & Hue Range & Saturation/Value Range \\
\hline
Color(Yellow) & $H \in (0.11, 0.22)$ & $S, V \in (0.4, 1.0)$ \\
\hline
\end{tabular}
\end{table}

Calculating the HSV value manually is discouraged due to the potential need for multiple attempts to accurately determine the specific color range. Therefore, \autoref{tab:hsv-yellow-color} establishes the range of path colors based on statistical analysis of HSV color pixel data. Furthermore, for edge detection for paths we have used the contour method, this method first analyzes the structure of binary images and follows the border. This extends the border following algorithm, which distinguishes between the outer borders and the internal hole borders of a binary image \cite{b23}.

For detecting path-centred points \cite{b24}\cite{b25} we have implemented Image moment and shape center calculation is done using spatial moments calculation of the contour with the following equation \eqref{eq:03}.

\begin{equation}
m_{ji} = x, y \cdot (x_i \cdot y_i)
\label{eq:03}
\end{equation}

The (x, y) array represents pixel values in the image, with 0 denoting black and 1 denoting white. Here, x corresponds to the row position, and y represents the column position within the image. Central moments are then computed as follows \eqref{eq:04}:

\begin{equation}
\mu_{ji} = x, y \cdot (x_i - x) \cdot (y_i - y)
\label{eq:04}
\end{equation}

where \((x, y)\) represents the center of mass.
\begin{align}
x &= \frac{m_{10}}{m_{00}} \\
y &= \frac{m_{01}}{m_{00}}
\end{align}

Contour moments are expressed similarly, but their calculation is based on Green's formula.

\begin{equation}
CL \, dx + M \, dy = \iint_D \left( \frac{\partial M}{\partial x} - \frac{\partial L}{\partial y} \right) \, dxdy
\end{equation}

Consider a contour C, which bounds a region D. If functions L and M are defined for points (x, y) within an open region encompassing D and possess continuous partial derivatives, and if the direction of integration along C is clockwise, then:

\subsubsection{Navigation and Measurement of Angle}
The angle between the reference vector and the central moments of the path is determined by using trigonometric functions. This angle can be represented as the function tan($\theta$), where $\theta$ signifies the angle.

\begin{equation}
\theta = \arctan\left(\frac{y}{x}\right)
\end{equation}

To convert the angle from radians to degrees, you can multiply the given angle by 180 and then divide it by $\pi$. This transformation is represented as:

Angle in degrees =( (Angle in radians × 180)/$\pi$ ) 

The transform range function is employed to further adjust the angle, as depicted in Equation (8), where values of x within the range [a, b] are transformed into values of y within the range [c, d].

\begin{equation}
y = x - ad - cb - a + c
\end{equation}

The angle calculation relies on the characteristics of the vehicle's wheel motors. A straightforward approach is suggested for fine-tuning the angle and steering the car using the equations (7) and (8). Below is the pseudocode for this method.

\begin{figure}[htbp]
\centering
\includegraphics[width=0.7\textwidth]{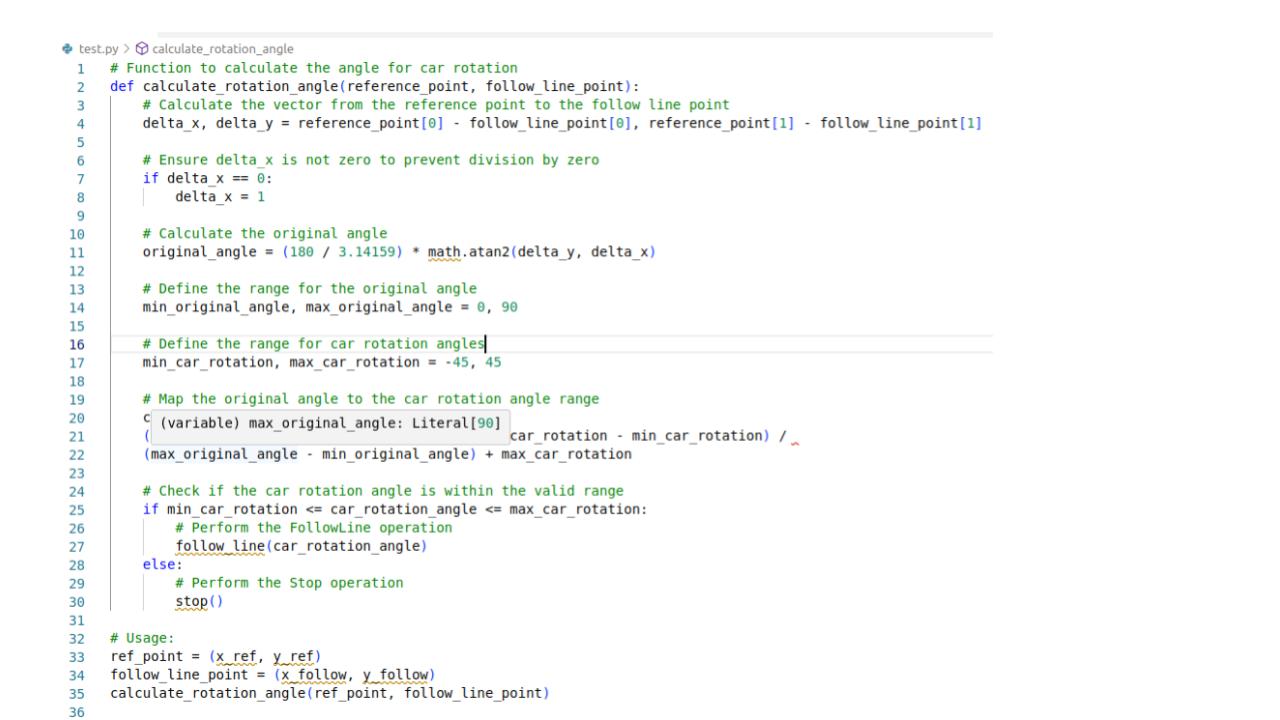} % Adjust the width as needed
\caption{Psudocode}
\label{pic06}
\end{figure}

This method \autoref{pic06} involves providing the reference vector and the origin of the path. Within the function, the initial angle on the unit circle is determined, which is then modified to align with the minimum and maximum rotation angles of the vehicle. Ultimately, by utilizing trigonometric functions and the transform range function, the rotation angle is computed for steering the vehicle.

\section{Experiment and Results}
\subsection{Experimental Configuration}
The proposed solution experiment was performed on the Ubuntu Operating System, and the system configuration contains 16 GB RAM, We have used VSCode with C++ release mode, and the communication method was performed on UART and CAN.

\subsection{Results}

\begin{figure}[htbp]
\centering
\includegraphics[width=0.4\textwidth]{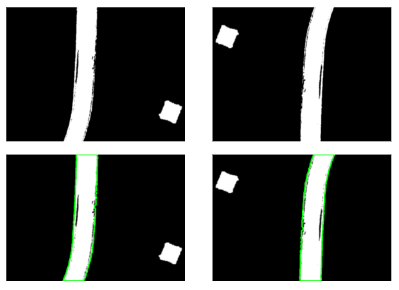} % Adjust the width as needed
\caption{(a) Contour detection is performed after applying an adaptive threshold, resulting in an image that includes an ambiguous object and a path. (b) Subsequent filtering of the contour image isolates the edges of the path.}
\label{pic07}
\end{figure}

In our proposed method AINS, the camera is fixed in front of the vehicle, so the path can be clearly visible. The actual path is detected using HSV and contour approach, but such input images have more than one contours, so first we remove noise and contours with the selection of the biggest or widest contours in the input image as described in \autoref{pic07}.

\begin{figure}[htbp]
\centering
\includegraphics[width=0.4\textwidth]{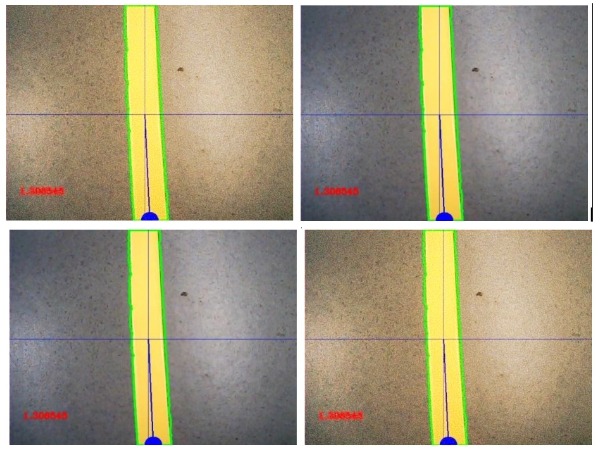} % Adjust the width as needed
\caption{An image illustrating the navigation method is shown, featuring the angle between the reference vector (indicated in blue text) and the centroid of the path (highlighted by a red semi-circle). The centroid of the path is depicted at the center of the crossed lines.}
\label{pic08}
\end{figure}

Once the contour is removed and the noise is reduced, we apply the angle calculation method to calculate the angle between vector and centroid points for the referenced path used by self-driving or autonomous cars. The prototype of the vision-based vehicle has been tested across a range of indoor environments. As a result, in particularly complex settings, some applications may experience minor impacts. However, these issues should be manageable as long as the HSV color range is appropriately calibrated, and the camera is correctly positioned. The ultimate outcome is illustrated in \autoref{pic08}.

Based on the experiments conducted, an alternative method for vehicle navigation, as opposed to utilizing a fixed angle, involves dividing the image into three regions: left, right, and center. In theory, using image region information seems like a plausible approach for navigation. However, this method presents a significant challenge. Maintaining a constant angle for vehicle navigation can be problematic since there's a high likelihood that the camera may not detect the path due to improper rotation. This underscores the need for dynamic angle calculations instead of relying solely on image region information.

A comparison between sensor-based and vision-based vehicles is provided in \autoref{tab:navigation-comparison}.

\begin{table}[htbp]
\centering
\caption{Sensor-Based vs. Vision-Based}
\label{tab:navigation-comparison}
\begin{tabularx}{\linewidth}{|X|X|X|X|}
\hline
Methods & Distance (m) & Estimated Error (\%) & Time Consumption (min:sec) \\
\hline
Sensor-Based & 100 & 4.6 & 3:51 \\
Vision-Based & 100 & 1.2 & 2:44 \\
\hline
\end{tabularx}
\end{table}

As evident in \autoref{tab:navigation-comparison}, the performance of the vision-based vehicle significantly outperforms that of the sensor-based vehicle. Both vehicles are operated at a constant speed of 22 kilometers per hour, and error calculations are made during vehicle maneuvers, considering factors such as rotation stability and environmental effects.

The vision-based approach proves to be more dependable compared to the sensor-based method. This is primarily because sensors require precise placement along the path, whereas a camera can detect the path as long as it remains within the camera's field of view.

In summary, our method entails detecting the vehicle's path through image processing techniques and subsequently calculating the centroid of the path. The angle between this centroid and the reference vector is then computed to steer the vehicle effectively.

\section{Discussion}
In summary, the AINS is a robust and cost-effective indoor navigation method for autonomous vehicles is proposed. The method. In the First step, the method is fed by an input video stream by a vehicle camera. In the second step, the path is detected after prepossessing the video. The method detects the path and turns based on the color and angle of the path. In the final step, the navigation is performed according to the path detected by the method. 

\section{Conclusion}

In recent times, researchers have been actively exploring innovative approaches to enhance the effectiveness and efficiency of autonomous vehicles, particularly in indoor settings. Autonomous vehicle navigation indoors presents numerous challenges, particularly due to the limited accuracy of GPS in such environments. While various robust methods have been examined to tackle this issue, many of these solutions suffer from high deployment costs.

In response to the challenges outlined above, we introduce an economical indoor navigation solution for autonomous vehicles called the Affordable Indoor Navigation Solution (AINS), which relies primarily on a monocular camera. Our proposed approach is centered around a single camera, eliminating the need for multiple, large, or power-intensive sensors like range finders and other navigation sensors. AINS demonstrates that it's possible to deploy indoor navigation systems for autonomous vehicles without incurring exorbitant costs.

Our method exhibits superior results compared to existing solutions, offering reductions in estimated error and time consumption

\section{Future Work}

We are planning to extend this work to robots enhancing the efficiency of algorithms with deep learning techniques so we can achieve better and higher proficiency in indoor scenarios, like deploying robots in libraries and other similar environment settings.

\vspace{12pt}
\color{red}
IEEE conference templates contain guidance text for composing and formatting conference papers. Please ensure that all template text is removed from your conference paper prior to submission to the conference. Failure to remove the template text from your paper may result in your paper not being published.


\begin{thebibliography}{00}
\bibitem{b1} Morales, Eduardo Sánchez, et al. "High precision indoor navigation for autonomous vehicles." 2019 International Conference on Indoor Positioning and Indoor Navigation (IPIN). IEEE, 2019.
\bibitem{b2} Hussein, A., Al-Kaff, A., de la Escalera, A., \& Armingol, J. M. (2015, November). Autonomous indoor navigation of low-cost quadcopters. In 2015 IEEE international conference on service operations and logistics, and informatics (SOLI) (pp. 133-138). IEEE.
\bibitem{b3} Kayalvizhi, S., et al. "A Comprehensive Study on Supermarket Indoor Navigation for Visually Impaired using Computer Vision Techniques." 2022 OPJU International Technology Conference on Emerging Technologies for Sustainable Development (OTCON). IEEE, 2023.
\bibitem{b4} Zang, S., Ding, M., Smith, D., Tyler, P., Rakotoarivelo, T., \& Kaafar, M. A. (2019). The impact of adverse weather conditions on autonomous vehicles: How rain, snow, fog, and hail affect the performance of a self-driving car. IEEE vehicular technology magazine, 14(2), 103-111.
\bibitem{b5} Bresson, Guillaume, et al. "Simultaneous localization and mapping: A survey of current trends in autonomous driving." IEEE Transactions on Intelligent Vehicles 2.3 (2017): 194-220.
\bibitem{b6} Cheng, J., Zhang, L., Chen, Q., Hu, X., \& Cai, J. (2022). A review of visual SLAM methods for autonomous driving vehicles. Engineering Applications of Artificial Intelligence, 114, 104992.
\bibitem{b7} Run, Ray-Shine, and Zhi-Yu Xiao. "Indoor autonomous vehicle navigation—a feasibility study based on infrared technology." Applied System Innovation 1.1 (2018): 4.
\bibitem{b8} Run, R. S., \& Xiao, Z. Y. (2018). Indoor autonomous vehicle navigation—a feasibility study based on infrared technology. Applied System Innovation, 1(1), 4.
\bibitem{b9} Chai, W., Chen, C., Edwan, E., Zhang, J., \& Loffeld, O. (2012, March). INS/Wi-Fi based indoor navigation using adaptive Kalman filtering and vehicle constraints. In 2012 9th Workshop on Positioning, Navigation and Communication (pp. 36-41). IEEE.
\bibitem{b10} Chiang, K. W., Huang, C. H., Chang, H. W., Lin, C. X., Tsai, M. L., Zeng, J. C., \& Hung, M. C. (2023). Semantic proximity update of GNSS/INS/VINS for Seamless Vehicular Navigation using Smartphone sensors. IEEE Internet of Things Journal.
\bibitem{b11} Cheng, J., Zhang, L., Chen, Q., Hu, X., \& Cai, J. (2022). A review of visual SLAM methods for autonomous driving vehicles. Engineering Applications of Artificial Intelligence, 114, 104992.
\bibitem{b12} A. D. Luca, M. Ferri, G. Oriolo and R. P. Giordano, “Visual Servoing with Exploitation of Redundancy: An Experimental Study,” in IEEE International Conference on Robotics and Automation (2008), pp. 3231–3237.
\bibitem{b13} Wu, Chih-Jen, and Wen-Hsiang Tsai. "Location estimation for indoor autonomous vehicle navigation by omni-directional vision using circular landmarks on ceilings." Robotics and Autonomous Systems 57.5 (2009): 546-555.
\bibitem{b14} Faragher, R. M., \& Harle, R. K. (2013, September). SmartSLAM-an efficient smartphone indoor positioning system exploiting machine learning and opportunistic sensing. In Proceedings of the 26th International Technical Meeting of The Satellite Division of the Institute of Navigation (ION GNSS+ 2013) (pp. 1006-1019).
\bibitem{b15} Ghofrani, A., Toroghi, R. M., \& Tabatabaie, S. M. (2020, January). ICPS-net: an end-to-end RGB-based indoor camera positioning system using deep convolutional neural networks. In Twelfth International Conference on Machine Vision (ICMV 2019) (Vol. 11433, pp. 572-578). SPIE.
\bibitem{b16} Chhikara, Prateek, et al. "DCNN-GA: A deep neural net architecture for navigation of UAV in indoor environment." IEEE Internet of Things Journal 8.6 (2020): 4448-4460.
\bibitem{b17} Celik, K., Chung, S. J., \& Somani, A. (2008, May). Mono-vision corner SLAM for indoor navigation. In 2008 IEEE International Conference on Electro/Information Technology (pp. 343-348). IEEE.
\bibitem{b18} Celik, Koray, and Arun K. Somani. "Monocular vision SLAM for indoor aerial vehicles." Journal of electrical and computer engineering 2013 (2013): 4-4.
\bibitem{b19} Atia, M. M., Liu, S., Nematallah, H., Karamat, T. B., \& Noureldin, A. (2015). Integrated indoor navigation system for ground vehicles with automatic 3-D alignment and position initialization. IEEE Transactions on Vehicular Technology, 64(4), 1279-1292.
\bibitem{b20} Gonzalez de Santos, L. M., Frias Nores, E., Martinez Sanchez, J., \& Gonzalez Jorge, H. (2021). Indoor path-planning algorithm for UAV-based contact inspection. Sensors, 21(2), 642.
\bibitem{b21} R. S. Run and Z. Y. Xiao, “Indoor Autonomous Vehicle Navigation—A Feasibility Study Based on Infrared Technology,” Appl. Syst. Innov. 1(1), 4–4 (2018).
\bibitem{b22} R. Nock and F. Nielsen, “Statistical Region Merging,” IEEE Trans. Pattern Anal. Mach. Intell. 26(11), 1452–1458 (2004).
\bibitem{b23} H. D. Cheng, X. H. Jiang, Y. Sun and J. Wang, “Color image segmentation: Advances and prospects,” Pattern Recognit. 34(12), 2259-2281 (2001).
\bibitem{b24} S. Suzuki and K. Abe, “Topological Structural Analysis of Digitized Binary Images by Border Following,” Comput. Vis. Graph. Image Process 30(1), 32-46 (1985).
\bibitem{b25} J. Flusser and T. Suk, “Rotation Moment Invariants for Recognition of Symmetric Objects,” IEEE Trans. Image Process. 15(12), 3784–3790 (2006).
\bibliographystyle{alpha}
\bibliography{sample}


\end{thebibliography}
\end{document}